\documentclass{article} 
\usepackage{iclr2026_conference_arxiv,times}

\makeatletter
\renewcommand{\@oddhead}{}    
\renewcommand{\@evenhead}{}   
\renewcommand{\@oddfoot}{\thepage}   
\renewcommand{\@evenfoot}{\thepage}
\makeatother

\usepackage{amsmath,amsfonts,bm}

\def\eqref#1{equation~\ref{#1}}

\def\1{\bm{1}}

\DeclareMathAlphabet{\mathsfit}{\encodingdefault}{\sfdefault}{m}{sl}
\SetMathAlphabet{\mathsfit}{bold}{\encodingdefault}{\sfdefault}{bx}{n}

\usepackage{graphicx}
\usepackage{url}
\usepackage{algorithm}
\usepackage{algorithmic}

\usepackage{array}
\usepackage{tabularx}
\usepackage{float}

\usepackage{amsmath}
\usepackage{graphicx}
\usepackage[colorlinks=true, allcolors=blue]{hyperref}
\usepackage{booktabs} 

\usepackage{amsmath}
\usepackage{algorithmic}
\usepackage{algorithm}
\usepackage{booktabs}
\usepackage{tabularx}
\usepackage{multirow}
\usepackage{rotating}
\usepackage{makecell, pifont}
\usepackage{threeparttable}
\usepackage{adjustbox}
\usepackage{ragged2e}

\usepackage{graphicx}
\usepackage{subcaption}   

\title{FlashHead: Efficient Drop-In Replacement for the Classification Head in Language Model Inference}

\author{%
  \makebox[\textwidth][c]{%
  \vspace{2cm}
    \begin{tabular}{c}
    \\
    Wilhelm Tranheden \quad
    Shahnawaz Ahmed \quad
    Devdatt Dubhashi \\
    Jonna Matthiesen \quad
    Hannes von Essen \\[0.5em]
    \textbf{Embedl} \\
    Gamla Almedalsv\"{a}gen 21, 412 63 Gothenburg, Sweden \\
    \texttt{\href{https://www.embedl.com}{https://www.embedl.com}}
    \end{tabular}
  }
}
\begin{document}
\pagestyle{fancy}
\fancyhf{}
\fancyfoot[C]{\thepage}
\renewcommand{\headrulewidth}{0pt}
\maketitle
\begin{abstract}
Language models are increasingly adopting smaller architectures optimized for consumer devices. In this setting, inference efficiency is the primary constraint. Meanwhile, vocabulary sizes continue to grow rapidly, making the classification head a critical bottleneck that accounts for up to 60\% of model parameters, and 50\% of inference compute. We introduce FlashHead, the first efficient drop-in replacement for the dense classification head that is training-free and hardware-friendly. FlashHead builds on principles from information retrieval, reframing that computation at the output head as a retrieval problem rather than a dense classification over the full vocabulary. FlashHead introduces four key innovations: (1) a balanced clustering scheme that structures vocabulary partitions into compact hardware-efficient tensors, (2) extending multiprobe retrieval to language model heads, enabling thousands of clusters to be scored in parallel, (3) a novel inference-time sampling mechanism that extends retrieval beyond top tokens, enabling probabilistic sampling across the full vocabulary, and (4) selective quantization, enabling effective low-bit computation in the head. Experiments on Llama-3.2, Gemma-3, and Qwen-3 show that FlashHead delivers model-level inference speedups of up to \textbf{1.75x} which maintaining output accuracy compared to the original head. By overcoming the classification head bottleneck, FlashHead establishes a new benchmark for efficient inference and removes a key barrier to developing smaller, capable models for consumer hardware.
\end{abstract}

%

\section{Introduction}


As small language models (SLMs) become central to wide adoption of machine learning systems across society, their deployment on consumer-grade hardware is becoming increasingly important \cite{Belcak2025_SLMAgenticAI, Shakhadri2024_Shakti, ACL2025_DemystifyingSLMs, EmpoweringEdgeAI2025}. In state-of-the-art SLM architectures such as Llama-3.2 \cite{llama3.2}, Gemma 3 \cite{gemma3}, and Qwen 3 \cite{yang2024qwen2}, the classification head alone accounts for between 20\% and 60\% of total model parameters and dominates inference cost. As the general trend is toward increasing vocabulary sizes \cite{tao2407scaling}, this problem is likely to be a central challenge for efficient inference in SLMs. 

Prior to the recent emphasis on SLMs, research on efficient deployment of large language models concentrated almost entirely on the transformer body. Quantization techniques ~\cite{frantar2022gptq, lee2024owq, lin2024give} were developed to compress weights and activations while maintaining model accuracy, yet these methods explicitly avoid quantizing the classification head. This limitation is also reflected in inference frameworks like vLLM~\cite{kwon2023efficient}, which as of the latest release leave the output head in higher precision to avoid accuracy degradation.

Prior work targeting the model head largely predates the current emphasis on SLMs, and relatively few approaches have been designed with inference efficiency as the primary goal. Proposed methods span hierarchical softmax~\cite{morin2005hierarchical} and vocabulary-trimming strategies~\cite{ushio2023efficient}, to adaptive output layers~\cite{zhao2021ann}. However, these typically require additional fine-tuning~\cite{NIPS2008_1e056d2b, morin2005hierarchical, joulin2017efficient, chen2015strategies, chen2025adaptive, zhao2021ann}, task-specific calibration~\cite{ushio2023efficient}, or fail to faithfully capture the full output probability distribution~\cite{shim2017svd, zhang2018navigating}.

At their core, these methods view the classification head as a two-stage retrieval mechanism. One family of approaches replaces the dense head with Approximate Nearest Neighbor (ANN) algorithms~\cite{zhang2018navigating, chen2025adaptive, zhao2021ann}. Variants of the inverted file (IVF) index are commonly used, and several methods~\cite{chen2025adaptive, zhao2021ann, chen2018learning} explicitly exploit spherical K-Means~\cite{dhillon2001concept} to align with the natural semantic structure of language models~\cite{petukhova2025text}. Despite their promise, these methods inherit the same drawbacks noted above.

Motivated by these limitations, we propose FlashHead: the first efficient drop-in replacement for the dense classification head that is both training-free and hardware-friendly. FlashHead introduces four key innovations:

\begin{enumerate}
\item {\bf Equal-sized clustering}: We employ a variant of spherical K-means that enforces strictly equal cluster sizes. This design allows centroids to be packed into compact tensors, enabling fast and predictable memory access. To our knowledge, FlashHead is the first LLM head that clusters token embeddings into strictly equal-sized spherical clusters.

\item {\bf Efficient multi-probing}:  Whereas prior work typically restricts retrieval to the single most likely cluster, our compact clustering enables aggressive use of multi-probing~\cite{lv2007multi}. We depart significantly from existing applications of information retrieval in language models by scaling from hundreds of clusters to tens of thousands. We achieve this by employing hundreds to thousands of probes to simultaneously score several clusters. This expanded probing strategy, designed specifically for inference in language models, is highly accurate while remaining efficient.

\item {\bf Probabilistic probe sampling}:  Unlike hierarchical softmax, which factorizes and learns the probability distribution end-to-end, our method naturally bridges retrieval and probabilistic decoding. We introduce a multinomial sampling step at the probe-selection stage, allowing top-token retrieval and sampling to be seamlessly unified.

\item {\bf Selective quantization}:  FlashHead admits coarse quantization in the first stage of its retrieval process. By distributing probability mass across such a large number of clusters and probes, it becomes naturally robust to quantization. This makes FlashHead the first method to enable safe and effective quantization of the output head. Flashhead in higher precision outperforms quantization of the dense head, and in combination with quantization delivers even greater performance.

\end{enumerate}

The four components are not independent optimizations but synergistic mechanisms, whose combination is crucial for the overall improvements, as demonstrated in our experiments. Overall, FlashHead turns the output head from a dominant bottleneck into a negligible component of inference. Crucially, the gains are not just at the head level: across all benchmarked models we see consistent \emph{model-level} speedups. For instance, FlashHead accelerates Llama-3.2-1B by up to 1.75×, Gemma-3-270M by 1.35×, and Qwen-3-1.7B by 1.31×, all without retraining and marginal accuracy loss. Even at the 8B scale, FlashHead delivers a 1.13× speedup where baseline inference already saturates consumer GPUs. In effect, what previously consumed a large portion of compute is compressed into an insignificant fraction, enabling state-of-the-art SLMs to run faster and leaner on commodity hardware.

Our main contributions can be summarized as follows:
\begin{enumerate}
\item We introduce FlashHead, the first training-free and hardware-friendly drop-in replacement for the dense classification head, which substantially improves inference efficiency while preserving accuracy.
\item We provide a comprehensive evaluation of FlashHead against existing alternatives, covering reasoning, multilingual, and instruction-following benchmarks across multiple model families, as well as detailed on-device latency measurements.
\item We release a public implementation of FlashHead integrated into several state-of-the-art models, including Llama 3.2, Gemma 3, and Qwen 3 (available at \textsc{[Redacted url due to blind review]}), to facilitate adoption and further research. 
\end{enumerate}

\section{Related Work}

Research on efficient alternatives to the dense classification head falls into two categories: trainable replacements and training-free replacements.


\subsection{Trainable replacements}

Several works replace the dense head with structures that must be retrained or fine-tuned alongside the model. Hierarchical softmax organizes the vocabulary into a tree to reduce complexity, but suffers from tree construction bias and limited flexibility ~\cite{morin2005hierarchical}. Adaptive softmax groups tokens by frequency and learns cluster heads jointly with the model ~\cite{joulin2017efficient}, while differentiated softmax allocates higher-rank embeddings to frequent tokens and lower-rank ones to rare tokens ~\cite{chen2015strategies}.

Prior work has also considered spherical K-means clustering of context vectors as a retrieval mechanism for inference ~\cite{chen2018learning}. These methods partition queries rather than keys and thus depend on training-time statistics to define candidate sets. ANN-based training methods such as ANN-Softmax ~\cite{chen2025adaptive} or MIDX Sampler ~\cite{zhao2021ann} adopt inverted-file or product-quantization indexes also based on K-means to accelerate training.

While several of these approaches rely on K-means to partition the vocabulary, they cluster queries (context vectors) and depend on training-time statistics or retraining to define candidate sets. FlashHead clusters the keys (token embeddings) once, offline, and introduces several innovations that enable a novel multi-probe retrieval, preserving the full probability distribution without any retraining or data.

\subsection{Training-free replacements}

Training-free replacements avoid retraining and therefore represent the most practical deployment alternatives. Vocabulary trimming ~\cite{ushio2023efficient} discards infrequent tokens using a calibration set, offering speedups but fundamentally restricting the vocabulary and reducing robustness to rare or out-of-distribution prompts. SVD-Softmax ~\cite{shim2017svd} pioneered low-rank decompositions for accelerating large-vocabulary softmax layers. However, because sampling occurs only after the coarse approximation step, probabilities are reliable primarily for high-likelihood tokens. Fast Graph Decoder ~\cite{zhang2018navigating} introduced approximate nearest neighbor search as a training-free replacement for classification heads. Its formulation, however, outputs only a top-$k$ candidate set rather than modeling the full probability distribution, limiting its applicability. With the exception of vocabulary trimming, these works were developed before the transformer era, when recurrent and embedding-based models dominated.

FlashHead belongs to the training-free category, while addressing existing limitations. In our experiments we therefore compare primarily to existing training-free replacements, which represent the most practical deployment alternatives.


\section{Methodology}
\label{dense-head}
In LLMs, the dense classification head transforms a $d$-dimensional hidden state vector $\mathbf{h} \in \mathbb{R}^{d}$ (where $d$ is the embedding size) to a vocabulary-sized logit vector. The hidden state vector encodes the contextual state produced by the model body after processing an input sequence. Concretely, the classification head applies a single matrix multiplication: $\mathbf{z} = \textbf{E}\times\mathbf{h}$, where $ \mathbf{E}\!\in\!\mathbb{R}^{v\times d}$ is the output‑embedding matrix (with $v$ being the total number of tokens in the vocabulary) to produce the logit vector. Each row $\mathbf{e}_i$ is the $d$‑dimensional embedding of token $i$.

For greedy retrieval the next token is selected as $t = argmax(\mathbf{z})$, i.e. the index of the largest logit in $z$.

For probabilistic token sampling, the logits are scaled by a temperature $\tau > 0$ and converted to a distribution
\[
\mathbf{y} \;=\; \operatorname{softmax}\!\Bigl(\tfrac{\mathbf{z}}{\tau}\Bigr),
\quad
t \;\sim\; \operatorname{Categorical}(\mathbf{y}).
\]


\subsection{Clustering the Embedding Matrix}
\label{clustering}

In FlashHead, a one-time offline clustering step is applied to partition the $v$ embedding vectors of $\mathbf{E}\!\in\!\mathbb{R}^{v\times d}$ into $c\!\ll\!v$ disjoint clusters
$\{\mathcal{C}_1,\dots,\mathcal{C}_c\}$. Specifically, \emph{spherical}
\textit{k}-means~\cite{dhillon2001concept}, a variant of
\textit{k}-means that measures similarity with the cosine metric, is applied. Cosine distance is a natural choice because empirical studies show that the semantic information in token embeddings is largely encoded in their \emph{direction}, with vector length playing only a minor role~\cite{mikolov2013distributed}. 


\paragraph{Objective.}
Given a number of clusters $c$, spherical \textit{k}-means minimizes the negative
cosine similarity between points and their assigned centroids~\cite{dhillon2001concept}:
\[
  \min_{\{\mathcal{C}_k\}}
  \;\sum_{k=1}^{c}\;\sum_{i\in\mathcal{C}_k}\;
  \bigl(1 - \mathbf{e}_i^{\!\top}\mathbf{c}_k \bigr),
  \qquad
  \text{with} \quad \mathbf{c}_k=\frac{\sum_{i\in\mathcal{C}_k}
  \mathbf{e}_i}{\bigl\lVert\sum_{i\in\mathcal{C}_k}
  \mathbf{e}_i\bigr\rVert_2}.
\]
Because every centroid is re‑normalised after each batch update, the algorithm
alternates between
(i)~assigning each token to the closest centroid on the sphere, and  
(ii)~recomputing centroid directions as the mean of the cluster members.

We implement a modified assignment step that enforces an equal
cluster size of $v/c$ tokens per cluster (thus $c$ is always a divisor of $v$). If a cluster exceeds its capacity, its lowest-similarity members are reassigned greedily to clusters that still have available slots, prioritizing the most similar eligible centroid. This ensures that all clusters have identical capacity while preserving semantic grouping. Enforcing \emph{exactly} equal sized clusters is critical for efficiency as discussed below.

\paragraph{Initialization and convergence.}
Empirically we converge within a budget of 1000 iterations for all models evaluated. Clustering is a one-time cost and goes relatively fast. For example clustering the 128\thinspace256-token vocabulary of
Llama‑3.2 into $c=8016$ clusters takes 4 hours on a single A40 GPU.

\paragraph{Centroid matrix.}
After clustering we store the normalized centroid matrix
$\mathbf{C}\!\in\!\mathbb{R}^{c\times d}$. In addition, we require the mapping
between clusters and token indices stored in a matrix
$\mathbf{C2T}\!\in\!\mathbb{R}^{c\times b}$, where $b$ is the size of the clusters. When clusters are not equalized (for ablation studies), see \ref{subsec:ablations}, it represents the largest cluster (rows corresponding to smaller clusters are padded).

\subsection{Token retrieval from Hidden Vector}
\label{sec:token-retrieval}
\begin{figure}[t]
  \centering
    \includegraphics[width=0.38\linewidth]{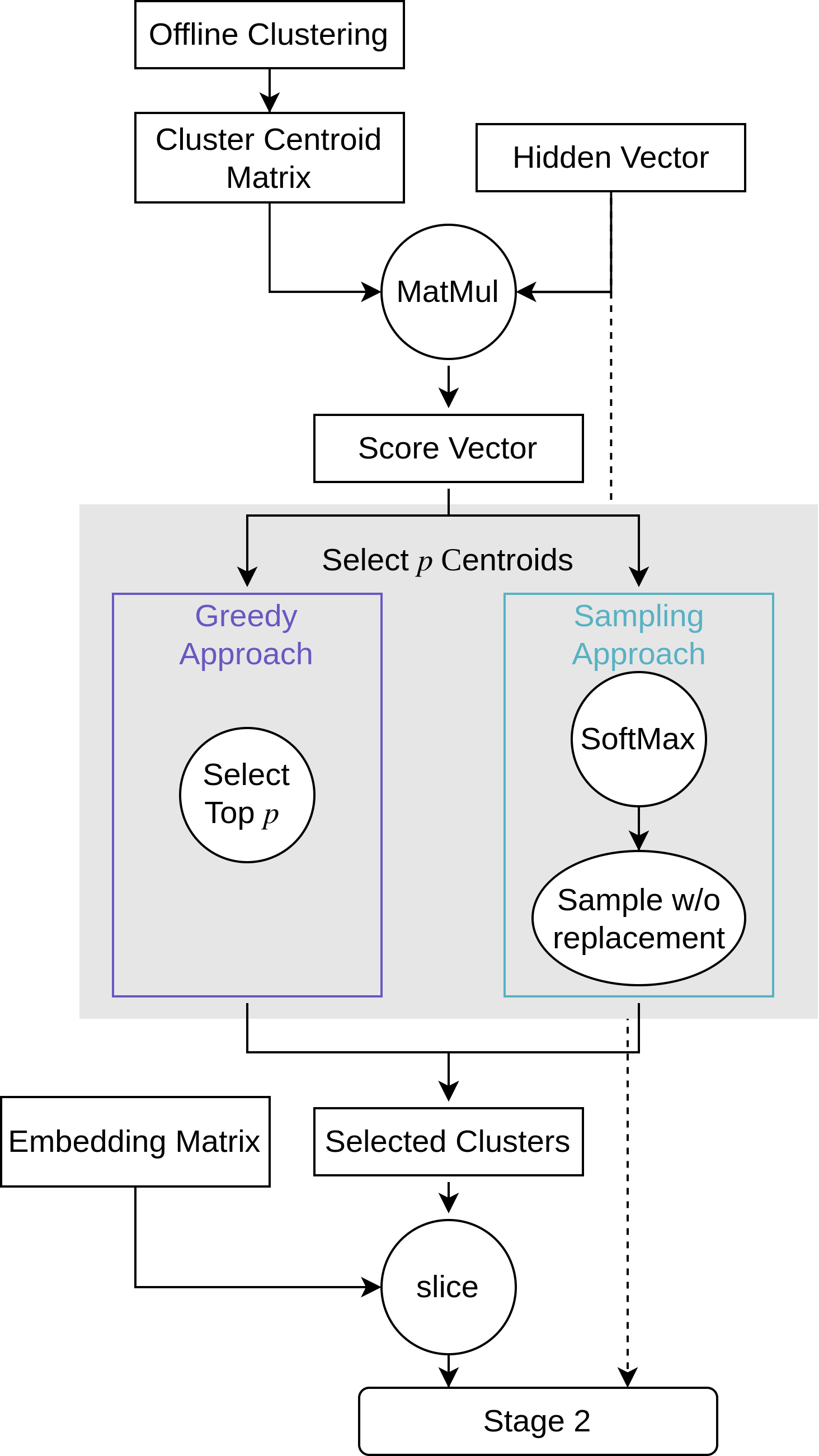}
    \includegraphics[width=0.38\linewidth]{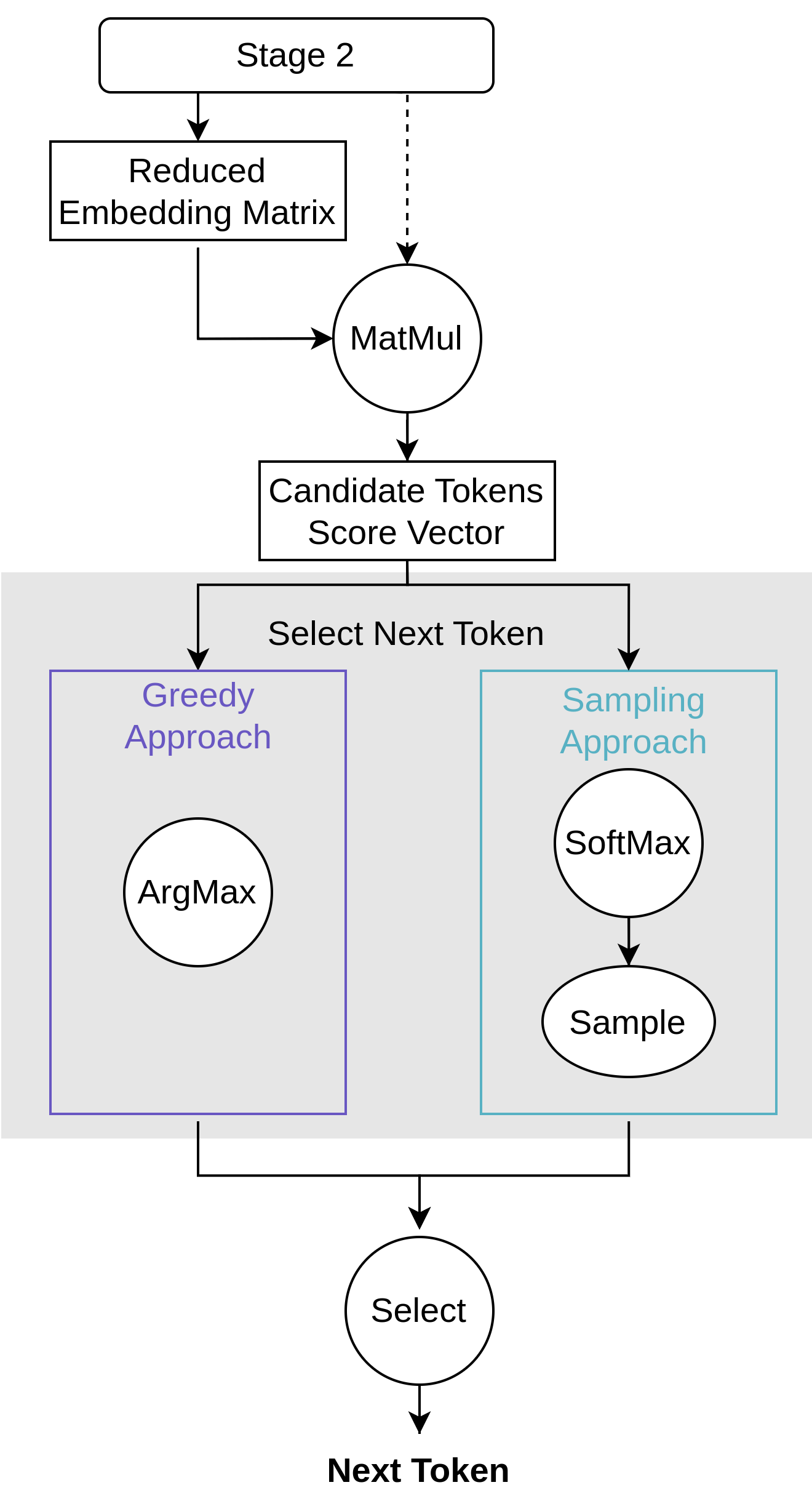}
  \caption{Several clusters of tokens are selected conditioned on the hidden vector produced by the model body. The reduced subset of token embeddings are multiplied with the hidden vector to make the final token selection.. Illustration of the proposed FlashHead algorithm, highlighting the difference between greedy approach of selecting the most likely next token, and the sampling approach allowing for probabilistic sampling of new tokens.}
  \label{fig:side_by_side}
\end{figure}

At inference time, FlashHead transforms the hidden state
$\mathbf{h}\!\in\!\mathbb{R}^{d}$ into the next token via a dynamic two-step retrieval process inspired by the inverted file (IVF) index (see Algorithm~\ref{alg:cap}). We support both decoding scenarios mentioned in Section~\ref{dense-head}; their key differences in FlashHead are illustrated in Figure~\ref{fig:side_by_side}.

\paragraph{Deterministic / greedy decoding.} For tasks such
as classification, after the matrix multiplication $\mathbf{C} \times \mathbf{h}$, the top $\mathit{p}$ centroids are selected given the produced centroid logits. FlashHead gathers tokens in these clusters into a reduced vocabulary embedding matrix $\tilde{\mathbf{E}}$. Given this embedding, token logits are computed over this smaller set, $\mathbf{z} = \tilde{\mathbf{E}} \times \mathbf{h}$, and the next token is chosen  as $t = argmax(\mathbf{z})$. 

\paragraph{Stochastic / sampling decoding.} For open‑ended generation, after the matrix multiplication $\mathbf{C} \times \mathbf{h}$, $p$ centroids are sampled without replacement, based on the softmax operation applied on the centroid logits. As for the deterministic/greedy decoding, FlashHead gathers tokens in these sampled clusters into a reduced vocabulary embedding matrix $\tilde{\mathbf{E}}$. In the second step, after the matrix multiplication $\tilde{\mathbf{E}} \times \mathbf{h}$, a token is sampled based on the softmax operation applied on the token logits.


\paragraph{Complexity.}
\label{par:complexiry}
The dense head requires $O(vd)$ multiplications, where $v$ is the vocabulary size.
FlashHead reduces this to
\[
O(cd + pbd),
\]
where $c$ is the number of clusters, $b = v/c$ the cluster size,
and $p \ll c$ the number of probes used at inference. In practice,
$p \cdot b \ll v$, yielding significant savings in both compute
and memory access. Notably, as $c$ increases, $b$ decreases, and our design choices allows a practical implementation where we can exploit this to scale to very large $c$ and $p$ while remaining efficient. We demonstrate this scalability and its impact on accuracy–efficiency trade-offs in Section~\ref{subsec:ablations}.

A critical source of efficiency comes from the cluster-to-token mapping $\mathbf{C2T}$. Although not reflected in the multiplication count above, $\mathbf{C2T}$ governs how token embeddings are retrieved after probe selection. Enforcing equal cluster sizes allows $\mathbf{C2T}$ to be stored
as a dense $[c \times b]$ tensor, so token indices can be computed with simple modular arithmetic. Without equal cluster sizes, $\mathbf{C2T}$ becomes ragged and requires per-cluster length tracking and masked gathers, which significantly increase memory access cost on accelerators. Section~\ref{subsec:ablations} shows that equal-sized clustering is essential to eliminate this overhead while preserving accuracy.

\paragraph{Quantization note.}
The first stage as described in Algorithm~\ref{alg:cap} (computing centroid logits $\mathbf{C} \times \mathbf{h}$) is
straightforward to quantize because (i)~ it is a fully static matrix multiplication,
and (ii)~ the coarse centroid probabilities are refined in the second stage using
higher-precision token logits. As shown in Section~\ref{subsec:ablations}, this
selective quantization yields significant speedups while maintaining accuracy.

\begin{algorithm}
    \caption{FlashHead.}\label{alg:cap}
    \begin{algorithmic}
        \item[\textbf{Input:}] Hidden vector generated by the model body $\mathbf{h} \in \mathbb{R}^{d}$, where $d$ is the embedding size, Embedding Matrix of all tokens $\mathbf{E} \in \mathbb{R}^{v \times d}$, where $v$ is the total number of tokens in the vocabulary.
        \item[\textbf{Hyperparameters:}] Number of probes $\mathit{p}$, Number of clusters $\mathit{k}$
        \item[\textbf{Output:}] Next token $\mathbf{t} \in \mathbb{R}^{d}$
        \item[\textbf{Initialize:}] cluster centroids $\mathbf{C} \in \mathbb{R}^{k \times d} \leftarrow$ cluster $\mathbf{E}$ into $k$ clusters
        \item[\textbf{Inference:}]
        \STATE $\mathbf{c'} \in \mathbb{R}^{k} \leftarrow \mathbf{C} \times \mathbf{h}$ \hfill \textit{(compute centroid logits)}
        \STATE $\mathbf{C}_{\mathit{p}} \in \mathbb{R}^{k \times p} \leftarrow$ select $\mathit{p}$ centroids given centroid logits $\mathbf{c'}$
        \STATE $\tilde{\mathbf{E}} \in \mathbb{R}^{x \times d} \leftarrow$ select tokens from $\mathbf{E}$ corresponding to selected centroids $\mathbf{C}_{\mathit{p}}$, according to $\mathbf{C2T}$.
        \STATE $\mathbf{z'} \in \mathbb{R}^{x} \leftarrow \tilde{\mathbf{E}} \times \mathbf{h}$ \hfill \textit{(compute token logits)}
        \STATE $t \in \mathbb{R}^{d} \leftarrow $ select next token based on token logits. \hfill \textit{}

    \end{algorithmic}
\end{algorithm}

\section{Experiments}
\label{sec:experiments}

Evaluating language models (LMs) is nontrivial because open‑ended generation intertwines correctness, fluency, and factuality~\cite{biderman2024lessons}; even small divergences in early tokens can cascade into valid but different continuations. We therefore report: (i)~standard LM benchmark results on generation and reasoning tasks; (ii)~\emph{Top‑$k$ Containment} to isolate head‑level fidelity; and (iii)~on‑device latency as time‑per‑output‑token, both for the head alone (\(\mathrm{TPOT}^{\mathrm{H}}\)) and the full model (\(\mathrm{TPOT}\)).

\paragraph{Models, datasets, and harness.}
We conduct extensive evaluations across several popular families of models: Llama‑3.2, Llama‑3.1, Qwen‑3, Gemma‑3. Unless stated, the base model is \texttt{Llama-3.2-1B-Instruct} with the LM‑Evaluation‑Harness for: MMLU‑Pro~\cite{wang2024mmlupro}, HellaSwag~\cite{zellers-etal-2019-hellaswag}, IFEval~\cite{zhou2023instruction}, BoolQ~\cite{clark2019boolq}, BBH~\cite{suzgun2022challenging}, TruthfulQA‑gen~\cite{lin-etal-2022-truthfulqa}, and GSM8K (8‑shot CoT). We use a combination of \texttt{alpaca\_eval}~\cite{alpaca2023,dubois2023alpacaeval}, \texttt{MATH‑Hard}~\cite{hendrycksmath2021} and XNLI~\cite{conneau2018xnli} to get a detailed head-level fidelity on challenging multilingual/logic datasets.
For probability‑based tasks (e.g., HellaSwag, BoolQ), FlashHead uses Monte Carlo sampling to estimate log‑likelihoods. To keep runtime tractable, we average over a 100‑example subset. We refer the reader to Appendix~\ref{app:mc} for more details on the Monte Carlo sampling. 

\paragraph{Baselines.}
We compare FlashHead to training‑free baselines: \emph{Vocabulary Trimming}~\cite{ushio2023efficient}, \emph{SVDSoftmax}~\cite{shim2017svd}, and \emph{Fast Graph Decoder (FGD)}~\cite{zhang2018navigating}. FGD is CPU‑only and float32 (as in the original implementation), so it is omitted where GPU or probability‑based scoring is required. For comparability, we keep method hyperparameters fixed across tasks: FlashHead uses \(c{=}8016\) clusters, \(p{=}512\) probes (unless varied in ablations or used for other models). Vocabulary Trimming keeps \(64{,}000\) tokens (50\% of the baseline vocabulary) while SVDSoftmax uses window size \(256\) (1/8 of embedding) and \(12{,}000\) top‑\(n\), both in line with original implementations. In case of FGD, we use a top $K$ of the nearest neighbors $384$, $efSearch=300$, and $M=40$, which are significantly larger than the values used in the original paper~\cite{zhang2018navigating} to match the larger vocabulary size for the baseline model.

\subsection{Common Evaluation Benchmarks}
\label{subsec:lm-eval-tasks}

Table~\ref{tab:main_results} summarizes task‑level performance. FlashHead is the only method to consistently \emph{match the baseline} on all aggregate metrics (within rounding). Notably, FlashHead accomplishes this feat while being significantly faster to execute on-device, see Section \ref{subsec:on-device}. In Appendix \ref{app:robust} we also provide the variance across these tasks between independent clustering runs.

\begin{table}[t]
  \caption{Evaluation on LM benchmarks. MMLU-Pro (exact\_match, custom-extract), HellaSwag (acc\_norm), IFEval (avg.\ of four metrics), BoolQ (acc\_norm), BBH (exact\_match), TruthfulQA-gen (bleu\_acc), GSM8K (exact\_match) for Llama 3.2 1B Instruct as the baseline.}
  \label{tab:main_results}
  \begin{threeparttable}
  \begin{tabularx}{\textwidth}{l *{7}{X}}
    \toprule
    Method & \shortstack{MMLU\\Pro} & \makecell{Hella\\Swag} & IFEval & BoolQ & BBH & \makecell{Truthful\\QA} & GSM8K \\
    \midrule
    Baseline         & 0.18 & 0.59 & 0.45 & 0.69 & 0.38 & 0.36 & 0.46 \\
    \midrule
    Vocab.\ Trimming & \textbf{0.18} & 0.53 & 0.35 & 0.65 & 0.37 & 0.36 & \textbf{0.46} \\
    SVDSoftmax       & 0.16 & 0.44 & 0.44 & 0.69 & 0.13 & 0.36 & 0.26 \\
    FGD              & —    & N/A  & 0.32 & N/A  & —    & \textbf{0.42} & — \\
    FlashHead & \textbf{0.18} & \textbf{0.59} & \textbf{0.45} & \textbf{0.69} & \textbf{0.38} & 0.36 & \textbf{0.46} \\
    \bottomrule
  \end{tabularx}
  \end{threeparttable}
\end{table}

\subsection{Top-K Containment}
\label{subsec:containment}

To isolate the classification head, we run both the baseline head and modified ones under identical hidden states and KV‑caches and compute \emph{Top‑$k$ Containment}: the fraction of cases where the token produced by a method appears in the baseline head’s top‑$k$. The results for FlashHead and other methods is aggregated in Table~\ref{tab:flashhead-benchmark-accuracy}. FlashHead alone attains a near-perfect match in top‑1 (within rounding) on all datasets except for the challenging multilingual dataset XNLI where it achieves \(97\%\). When looking at top-3, FlashHead alone achieves a near-perfect match even for XNLI.

\begin{table}[t]
  \caption{Top‑$k$ Containment (\(c{=}8016\), \(p{=}512\)) vs.\ training‑free baselines on \texttt{Llama‑3.2‑1B‑Instruct}. For \(k{=}1\), containment = accuracy.}
  \label{tab:flashhead-benchmark-accuracy}
  \centering
  \begin{tabular}{lcccccc}
    \toprule
    \multirow{2}{*}{Method} & \multicolumn{3}{c}{Top‑1} & \multicolumn{3}{c}{Top‑3}\\
     & Alpaca & MATH‑Hard & XNLI  & Alpaca & MATH‑Hard & XNLI \\
    \midrule
     Baseline         & 1.00 & 1.00 & 1.00 & 1.00 & 1.00 & 1.00 \\
    \midrule
     Vocab.\ Trimming & 0.99 & 0.99 & 0.51 & \textbf{1.00} & \textbf{1.00} & 0.69 \\
     SVDSoftmax       & 0.94 & 0.96 & 0.96 & 0.99 & 0.99 & 0.99 \\
     FlashHead     & \textbf{1.00} & \textbf{1.00} & \textbf{0.97} & \textbf{1.00} & \textbf{1.00} & \textbf{1.00} \\
    \bottomrule
  \end{tabular}
\end{table}

\subsection{On‑device Latency}
\label{subsec:on-device}

We report \(\mathrm{TPOT}^{\mathrm{H}}\) (head‑only) and end‑to‑end (model-level) \(\mathrm{TPOT}\) on an NVIDIA RTX 3500 Ada Generation GPU; additional CPU results and effective parameter reductions are deferred to Appendix~\ref{app: on-device} due to space. Unless noted otherwise, latencies are measured using vLLM~\cite{kwon2023efficient}, averaged over 128 generated tokens, averaged over 100 prompts, with 10 warm-up prompts, in either bfloat16 or W4A16 (int4) precisions for model computations.

Unless otherwise stated, the batch-size is 1. Because our focus is edge-inference and consumer devices, where both memory is limited and user queries must be answered immediately, batching is typically not applicable. As prior work shows~\cite{jiang2018efficient, wofk2019fastdepth, lin2022device}), latency and throughput are commonly reported at batch size 1 to reflect realistic on-device conditions.

Table~\ref{tab:gpu-latency} shows that FlashHead achieves the best \(\mathrm{TPOT}^{\mathrm{H}}\) and significantly improves full‑model \(\mathrm{TPOT}\). 

\begin{table}[t]
  \caption{GPU latency (ms) for head (\(\mathrm{TPOT}^{\mathrm{H}}\)) and full model (\(\mathrm{TPOT}\)) under bfloat16 and int4 heads. Note that FGD lacks GPU support. Extended CPU and parameter results appear in the Appendix~\ref{app: on-device}}
  \label{tab:gpu-latency}
  \centering
  \begin{tabular}{lccc}
    \toprule
    Method & TPOT$^\text{H}$ $\downarrow$ & TPOT (BF16) $\downarrow$ & TPOT (INT4) $\downarrow$ \\
    \midrule
    Baseline          & 1.94 & 7.69 & 3.60 \\
    Vocab.\ Trimming  & 1.07 (1.81$\times$) & 6.82 (1.13$\times$) & 2.73 (1.32$\times$) \\
    SVDSoftmax        & 0.61 (3.18$\times$) & 6.36 (1.21$\times$) & 2.27 (1.59$\times$) \\
    FGD               & N/A & N/A & N/A \\
    FlashHead      & \textbf{0.40 (4.85$\times$)} & \textbf{6.15 (1.25$\times$)} & \textbf{2.06 (1.75$\times$)} \\
    \bottomrule
  \end{tabular}
\end{table}

\subsection{Generalization across models}
Table~\ref{tab:flashhead-main} demonstrates that the head‑level speedups translate across families (Llama‑3.2, Llama‑3.1, Qwen‑3, Gemma‑3). These results highlight FlashHead's robustness as a drop‑in head for a wide variety of language models spanning parameter ranges from 270 million parameters to 8 billion.
\begin{table}[t]
\setlength{\tabcolsep}{3pt}
  \caption{BBH and GPU latency improvements across models. Values in parentheses are speedup factors vs.\ the dense head. A comprehensive accuracy table and CPU metrics are included in Appendix~\ref{app: more-models}. Note that the 8 billion parameter Llama 3.1 model is too big for inference in bfloat16 precision (Out-of-Memory, OOM).}
  \label{tab:flashhead-main}
  \centering
  \begin{tabular}{lcccc}
    \toprule
    Model & BBH $\uparrow$ & TPOT$^\text{H}$ $\downarrow$ & TPOT (BF16) $\downarrow$ & TPOT (INT4) $\downarrow$ \\
    \midrule
    Llama‑3.2‑1B  & 0.38 $\rightarrow$ 0.38 & 1.94 $\rightarrow$ 0.40 (4.9$\times$) & 7.69 $\rightarrow$ 6.15 (1.2$\times$) & 3.60 $\rightarrow$ 2.06 (1.8$\times$) \\
    Llama‑3.2‑3B  & 0.57 $\rightarrow$ 0.57 & 2.13 $\rightarrow$ 0.68 (3.1$\times$) & 18.60 $\rightarrow$ 17.15 (1.1$\times$) & 7.11 $\rightarrow$ 5.66 (1.3$\times$) \\
    Llama‑3.1‑8B  & 0.71 $\rightarrow$ 0.70 & 2.72 $\rightarrow$ 1.21 (2.2$\times$) & OOM & 13.55 $\rightarrow$ 12.04 (1.1$\times$) \\
    Qwen‑3‑1.7B   & 0.45 $\rightarrow$ 0.45 & 1.61 $\rightarrow$ 0.45 (3.6$\times$) & 9.97 $\rightarrow$ 8.81 (1.1$\times$) & 4.85 $\rightarrow$ 3.69 (1.3$\times$) \\
    Gemma‑3‑270M  & 0.27 $\rightarrow$ 0.27 & 0.99 $\rightarrow$ 0.37 (2.7$\times$) & 2.52 $\rightarrow$ 1.90 (1.3$\times$) & 2.38 $\rightarrow$ 1.76 (1.4$\times$) \\
    Gemma‑3‑1B    & 0.38 $\rightarrow$ 0.38 & 1.66 $\rightarrow$ 0.52 (3.2$\times$) & 6.77 $\rightarrow$ 5.63 (1.2$\times$) & 4.12 $\rightarrow$ 2.98 (1.4$\times$) \\
    \bottomrule
  \end{tabular}
\end{table}

\subsection{Ablations}
\label{subsec:ablations}

\paragraph{Quantization of the head.}
Quantizing dense LM heads often degrades accuracy, which is why deployment stacks typically avoid it. In contrast, FlashHead’s stage‑1 (centroid scoring) is robust to low‑bit quantization because final token probabilities are re-generated in stage‑2 in higher precision. Table~\ref{tab:flashhead-quantization} shows that FlashHead with an int4 stage‑1 achieves \(\mathrm{TPOT}^{\mathrm{H}}\) gains over a quantized dense head while preserving BBH accuracy much better (bfloat16 FlashHead already matches the baseline).

\begin{table}[t]
  \caption{Impact of head quantization on BBH and GPU \(\mathrm{TPOT}^{\mathrm{H}}\). Latencies are measured and quantization is performed with the transformers~\cite{wolf2020huggingfacestransformersstateoftheartnatural} and HQQ~\cite{badri2023hqq} libraries, due to their support of quantizing language model heads.}
  \label{tab:flashhead-quantization}
  \centering
  \begin{tabular}{llcc}
    \toprule
    Method & Precision & BBH $\uparrow$ & GPU TPOT$^\text{H}$ $\downarrow$ \\
    \midrule
    Baseline   & BF16 & 0.381 & 1.433 \\
               & INT4 & 0.369 & 0.486 \\
    \midrule
    FlashHead & BF16 & \textbf{0.381} & \textbf{0.320} \\
               & INT4 & \textbf{0.379} & \textbf{0.258} \\
    \bottomrule
  \end{tabular}
\end{table}

\paragraph{Equal-sized clustering.}
Equal‑size clusters enable a dense cluster‑to‑token mapping (\(C2T\)) with predictable memory access and lower gather overhead on accelerators. Table~\ref{tab:flashhead-balanced-clustering} quantifies the benefit: equal-sized clustering simultaneously \emph{improves accuracy and reduces} \(\mathrm{TPOT}^{\mathrm{H}}\).

\begin{table}[t]
  \caption{Equal vs.\ Unequal clustering on BBH and GPU \(\mathrm{TPOT}^{\mathrm{H}}\).}
  \label{tab:flashhead-balanced-clustering}
  \centering
  \begin{tabular}{lcc}
    \toprule
    Method & BBH $\uparrow$ & GPU TPOT$^\text{H}$ $\downarrow$ \\
    \midrule
    Unequal clusters & 0.371 & 0.520 \\
    Equal clusters   & \textbf{0.381} & \textbf{0.320} \\
    \bottomrule
  \end{tabular}
\end{table}

\paragraph{Number of clusters and probes.}
FlashHead exposes a smooth latency–accuracy trade‑off: increasing probes~\(p\) or clusters~\(c\) improves containment, with modest effect on \(\mathrm{TPOT}\) because cluster size \(b{=}v/c\) shrinks as \(c\) grows. Table~\ref{tab:llama-flashhead-latency} shows that while these parameters can be used to trade accuracy for further latency gains, we pick \((c{=}8016,p{=}512)\) as the default to showcase FlashHead’s unique ability to maintain accuracy while delivering significant model-level speedups.

\begin{table}[H]
\setlength{\tabcolsep}{3pt}
  \caption{Top‑1 containment and GPU latency as a function of clusters \(c\) and probes \(p\). We report head latency (\(\mathrm{TPOT}^{\mathrm{H}}\)) and full‑model \(\mathrm{TPOT}\) in BF16/INT4.}
  \label{tab:llama-flashhead-latency}
  \centering
  \begin{tabular}{llcccccc}
    \toprule
    \multirow{2}{*}{\#clusters} & \multirow{2}{*}{\#probes} 
      & \multicolumn{3}{c}{Top‑1} 
      & \multicolumn{3}{c}{GPU Latency (ms)} \\
    \cmidrule(lr){3-5} \cmidrule(lr){6-8}
     &  & Alpaca & MATH‑Hard & XNLI  
        & TPOT$^\text{H}$ $\downarrow$ 
        & \makecell{TPOT\\(BF16) $\downarrow$} 
        & \makecell{TPOT\\(INT4) $\downarrow$} \\
    \midrule 
    \multicolumn{2}{c}{Baseline} & 1.00 & 1.00 & 1.00 & 1.94 & 7.69 & 3.60 \\
    \midrule 
    4008 & 128  & 0.99 & 0.99 & 0.93 & 0.18 (10.78$\times$) & 5.93 (1.30$\times$) & 1.84 (1.96$\times$) \\
         & 256  & 0.99 & 0.99 & 0.95 & 0.33 (5.88$\times$)  & 6.08 (1.26$\times$) & 1.99 (1.80$\times$) \\
         & 512  & 1.00 & 1.00 & 0.97 & 0.58 (3.34$\times$)  & 6.33 (1.21$\times$) & 2.24 (1.61$\times$) \\
    \midrule
    8016 & 128  & 0.98 & 0.99 & 0.93 & 0.12 (16.17$\times$) & 5.87 (1.31$\times$) & 1.78 (2.02$\times$) \\
         & 256  & 0.99 & 1.00 & 0.96 & 0.17 (11.41$\times$) & 5.92 (1.30$\times$) & 1.83 (1.97$\times$) \\ 
         & 512  & 1.00 & 1.00 & 0.97 & 0.40 (4.85$\times$)  & 6.15 (1.25$\times$) & 2.06 (1.75$\times$) \\ 
    \midrule
    16032& 128  & 0.99 & 1.00 & 0.96 & 0.21 (9.24$\times$)  & 5.96 (1.29$\times$) & 1.87 (1.92$\times$) \\ 
         & 256  & 0.99 & 0.99 & 0.97 & 0.28 (6.93$\times$)  & 6.03 (1.28$\times$) & 1.94 (1.86$\times$) \\
         & 512  & 1.00 & 1.00 & 0.99 & 0.30 (6.47$\times$)  & 6.05 (1.27$\times$) & 1.96 (1.84$\times$) \\
    \bottomrule
  \end{tabular}
\end{table}

\section{Discussion}
\label{sec:discussion}

FlashHead is most impactful for small and mid-scale language models, where the classification head constitutes a large fraction of parameters and runtime. In such settings, our results indicate that FlashHead reduces the head to a negligible share of overall cost, yielding significant model-level speedups without any fine-tuning. 

A promising direction is extending FlashHead to server-side deployment scenarios where large batch sizes are critical. In particular, speculative decoding \cite{zhao2025fr} has emerged as a key technique to accelerate inference by generating multiple candidate tokens in parallel. This line of work would adapt FlashHead beyond edge inference toward high-throughput server settings, where latency–throughput trade-offs differ substantially from consumer hardware. Future work may also explore tighter integration with inference frameworks, low-level kernel optimizations, and extensions beyond text, such as adapting FlashHead to multimodal models. Another promising direction is to combine FlashHead with training-time modifications, or to accelerate the training process itself, potentially improving efficiency further.

There are also limitations. FlashHead does not provide a closed-form probability distribution over the entire vocabulary. This is not a drawback at inference time, but for evaluation in a research setting, we need to compute probability distributions over all tokens in the vocabulary which we currently do via Monte Carlo simulations, see Appendix~\ref{app:mc} for more details, which slows down likelihood-based evaluation. The clustering itself introduces modest additional storage, though this overhead is negligible compared to full model size.

\section{Conclusions}
We introduced FlashHead, the first training-free and hardware-friendly replacement for the dense classification head in language models. By combining equal-sized clustering, aggressive multi-probe retrieval, probabilistic probe sampling, and selective quantization, FlashHead collapses the head from a dominant bottleneck into a negligible cost. Across Llama-3.2, Llama 3.1, Gemma-3, and Qwen-3, it preserves accuracy while delivering up to 1.75× end-to-end speedups, even at billion-parameter scales. By removing a key obstacle to deployment on consumer hardware, FlashHead sets a new baseline for output-layer design and opens the path toward faster, smaller, and more widely usable language models.



\section{Reproducibility statement}

We have made all efforts to ensure that the results presented in this paper can be independently reproduced. The FlashHead algorithm is clearly outlined as pseudocode, allowing readers to understand and re-implement it without ambiguity. All hyperparameters and experimental settings are reported in detail in the respective tables to enable faithful reproduction of the experiments. Furthermore, we provide code implementations of FlashHead for several models. We believe these steps will enable the community to readily reproduce, verify, and build upon our work.
\bibliography{references}
\bibliographystyle{iclr2026_conference}

\clearpage
\appendix
\section{Appendix: Expanded On-device Benchmarks}
\label{app: on-device}
In the main text (Section~\ref{subsec:on-device}), we presented GPU results for FlashHead and baselines, demonstrating that FlashHead consistently reduces both head-only and full-model latency.  
Here, we expand those results with additional \textbf{CPU latencies} and \textbf{active parameter counts} for completeness.  The CPU device is Intel Core i7-10870H @
242 2.20GHz × 16 CPU, and measurements are performed using the transformers~\cite{wolf2020huggingfacestransformersstateoftheartnatural} and HQQ~\cite{badri2023hqq} libraries.

These results confirm that FlashHead delivers speedups across both CPU and GPU, while reducing the number of active parameters in the head to a fraction of the dense baseline.

\begin{table}[H]
  \caption{On-device latencies and parameter counts for all methods. 
  Latencies are reported in time-per-output-token (ms), averaged over 128 generated tokens. 
  Factors in parentheses indicate relative speedup or reduction compared to the baseline.}
  \label{tab:flashhead-benchmark-latency}
  \centering

  \subfloat[GPU Latency]{
  \begin{tabular}{lccc}
    \toprule
    Method & TPOT$^\text{H}$ $\downarrow$ & TPOT (BF16) $\downarrow$ & TPOT (INT4) $\downarrow$ \\
    \midrule
    Baseline        & 1.94 & 7.69 & 3.60 \\
    Vocab.\ Trimming & 1.07 (1.81$\times$) & 6.82 (1.13$\times$) & 2.73 (1.32$\times$) \\
    SVDSoftmax      & 0.61 (3.18$\times$) & 6.36 (1.21$\times$) & 2.27 (1.59$\times$) \\
    FGD             & N/A & N/A & N/A \\
    FlashHead & \textbf{0.40 (4.85$\times$)} & \textbf{6.15 (1.25$\times$)} & \textbf{2.06 (1.75$\times$)} \\
    \bottomrule
  \end{tabular}
  }

  \vspace{0.8em}

  \subfloat[CPU Latency (BF16)]{
  \begin{tabular}{lcc}
    \toprule
    Method & TPOT$^\text{H}$ $\downarrow$ & TPOT $\downarrow$ \\
    \midrule
    Baseline        & 15.92 & 85.75 \\
    Vocab.\ Trimming & 7.74 (2.06$\times$) & 77.56 (1.11$\times$) \\
    SVDSoftmax      & 30.94 (0.51$\times$) & 100.77 (0.85$\times$) \\
    FGD             & 4.74 (3.36$\times$) & 74.56 (1.15$\times$) \\
    FlashHead & \textbf{3.73 (4.27$\times$)} & \textbf{73.55 (1.17$\times$)} \\
    \bottomrule
  \end{tabular}
  }

  \vspace{0.8em}

  \subfloat[Active Parameters]{
  \begin{tabular}{lcc}
    \toprule
    Method & Params$^\text{H}$ $\downarrow$ & Params $\downarrow$ \\
    \midrule
    Baseline        & 263M & 1.236B \\
    Vocab.\ Trimming & 131M & 1.104B \\
    SVDSoftmax      & 59M & 1.032B \\
    FGD             & \textbf{0.79M} & \textbf{974M} \\
    FlashHead & 33M & 1.006B \\
    \bottomrule
  \end{tabular}
  }

\end{table}

\section{Appendix: Expanded Results Across Models}
\label{app: more-models}
In Section~\ref{sec:experiments}, we showed that FlashHead generalizes across model families and scales, providing consistent speedups from small (270M) to large (8B) models.  
Here, we provide a more comprehensive view with two sets of results:  
(a)~benchmark accuracy across LM tasks, and  
(b)~efficiency metrics including head parameters, total parameters, and latency.  
These expanded results confirm that FlashHead’s gains are robust across model families.

\begin{table}[H]
  \setlength{\tabcolsep}{8.5pt} 
  \centering
  \caption{Evaluation of models across (a) LM benchmarks and (b) efficiency metrics. We exclude BoolQ and HellaSwag due to the high cost of Monte Carlo simulations. Params$^\text{H}$ and Params indicate active and total parameters.}
  \label{tab:lm_and_tpot}
  
  \subfloat[LM evaluation benchmarks]{
    \begin{tabular}{l l c c c c c}
      \toprule
      Model & Method & \makecell{MMLU-\\Pro} & BBH & \makecell{Truthful\\QA} & IFEval & GSM8k \\
      \midrule
      \multirow{2}{*}{Llama-3.2-3B} & Baseline  & 0.31 & 0.57 & 0.57 & 0.57 & 0.77 \\
                                    & FlashHead & 0.31 & 0.57 & 0.58 & 0.56 & 0.77 \\
      \cmidrule(lr){1-7}
      \multirow{2}{*}{Llama-3.1-8B} & Baseline  & 0.41 & 0.71 & 0.62 & 0.53 & 0.85 \\
                                    & FlashHead & 0.41 & 0.70 & 0.62 & 0.52 & 0.85 \\
      \cmidrule(lr){1-7}
      \multirow{2}{*}{Llama-3.2-1B} & Baseline  & 0.18 & 0.38 & 0.36 & 0.45 & 0.47 \\
                                    & FlashHead & 0.18 & 0.38 & 0.36 & 0.45 & 0.46 \\
      \cmidrule(lr){1-7}
      \multirow{2}{*}{Qwen-3-1.7B}  & Baseline  & 0.38 & 0.45 & 0.47 & 0.24 & 0.13 \\
                                    & FlashHead & 0.38 & 0.45 & 0.47 & 0.25 & 0.12 \\
      \cmidrule(lr){1-7}
      \multirow{2}{*}{Gemma-3-1B}   & Baseline  & 0.15 & 0.38 & 0.31 & 0.55 & 0.42 \\
                                    & FlashHead & 0.15 & 0.38 & 0.31 & 0.49 & 0.39 \\
      \cmidrule(lr){1-7}
      \multirow{2}{*}{Gemma-3-270M} & Baseline  & 0.09 & 0.27 & 0.31 & 0.32 & 0.02 \\
                                    & FlashHead & 0.09 & 0.27 & 0.32 & 0.30 & 0.02 \\
      \bottomrule
    \end{tabular}
  }
  
  \vspace{0.8em}
  
  \subfloat[Efficiency and throughput metrics]{
  \setlength{\tabcolsep}{2pt} 
    \begin{tabular}{l l c c c c c}
      \toprule
      Model & Method 
            & Params$^\text{H}$ $\downarrow$ 
            & Params $\downarrow$ 
            & TPOT$^\text{H}$ $\downarrow$ 
            & \makecell{TPOT\\(BF16) $\downarrow$} 
            & \makecell{TPOT\\(INT4) $\downarrow$} \\
      \midrule
      \multirow{2}{*}{Llama-3.2-1B}   & Baseline  & 262M & 1.236B & 1.94 & 7.69 & 3.60 \\
                                      & FlashHead & 33M  & 1.006B & 0.40 (4.85$\times$) & 6.15 (1.25$\times$) & 2.06 (1.75$\times$) \\
      \cmidrule(lr){1-7}
      \multirow{2}{*}{Llama-3.2-3B}   & Baseline  & 394M & 3.213B & 2.13 & 18.60 & 7.11 \\
                                      & FlashHead & 50M  & 2.869B & 0.68 (3.13$\times$) & 17.15 (1.08$\times$) & 5.66 (1.26$\times$) \\
      \cmidrule(lr){1-7}
      \multirow{2}{*}{Llama-3.1-8B}   & Baseline  & 525M & 8.030B & 2.72 & OOM & 13.55 \\
                                      & FlashHead & 100M & 7.605B & 1.21 (2.25$\times$) & OOM & 12.04 (1.13$\times$) \\
      \cmidrule(lr){1-7}
      \multirow{2}{*}{Qwen-3-1.7B}    & Baseline  & 311M & 1.721B & 1.61 & 9.97 & 4.85 \\
                                      & FlashHead & 36M  & 1.446B & 0.45 (3.58$\times$) & 8.81 (1.13$\times$) & 3.69 (1.31$\times$) \\
      \cmidrule(lr){1-7}
      \multirow{2}{*}{Gemma-3-270M}   & Baseline  & 168M & 268M  & 0.99 & 2.52 & 2.38 \\
                                      & FlashHead & 21M  & 121M  & 0.37 (2.68$\times$) & 1.90 (1.33$\times$) & 1.76 (1.35$\times$) \\
      \cmidrule(lr){1-7}
      \multirow{2}{*}{Gemma-3-1B}     & Baseline  & 302M & 1.000B & 1.66 & 6.77 & 4.12 \\
                                      & FlashHead & 38M  & 736M  & 0.52 (3.19$\times$) & 5.63 (1.20$\times$) & 2.98 (1.38$\times$) \\
      \bottomrule
    \end{tabular}
  }
\end{table}

\section{Robustness to independent Clustering runs}
\label{app:robust}
FlashHead relies on a one-time offline spherical $k$-means clustering of token embeddings. Because $k$-means is sensitive to random initializations, it is natural to ask whether rebuilding the clusters from scratch affects
either predictive quality.

To answer this question, we repeat the common evaluation benchmarks used for the main FlashHead results reported in the main text \emph{five (5) times}, regenerating the complete set of $8\,016$ clusters in each run.

Table~\ref{tab:lmeval_variance_flashhead} reports the mean and
standard deviation across the five runs.  Rebuilding the clustering neither helps nor hurts any benchmark.

\begin{table}[H]
    \caption{Mean and standard deviation of FlashHead performance on LM-Eval benchmarks (Llama-3.2-1B-Instruct); scores
  are averaged over five independent clusterings.}
    \label{tab:lmeval_variance_flashhead}
    \begin{tabularx}{\textwidth}{*{5}{>{\raggedleft\arraybackslash}X}}
      \toprule
      MMLU-Pro & BBH & TruthfulQA (gen) & IFEval & GSM8K (chat) \\
      \midrule
      0.181 $\pm$ 0.001 & 0.377 $\pm$ 0.004 & 0.363 $\pm$ 0.002 & 0.452 $\pm$ 0.003 & 0.465 $\pm$ 0.003 \\
      \bottomrule
    \end{tabularx}
  \end{table}

\section{Estimating the Marginal Distribution}

For evaluation tasks such as HellaSwag and BoolQ,
we need log-likelihoods w.r.t.\ the \emph{full} vocabulary, but enumerating all $c$ clusters is precisely what FlashHead avoids.

FlashHead first samples a \emph{subset} of centroids
$S\subseteq \mathcal{C} := \{\mathcal{C}_1,\dots,\mathcal{C}_c\}$ (size $n_{\text{probes}}$) and then samples a
token $t$ within those centroids, see Figure \ref{fig:side_by_side}. Consequently the exact marginal
$
  p(t\mid\mathbf{h})
  =\mathbb{E}_{S}\bigl[p(t\mid\mathbf{h},S)\bigr]
$
requires an expectation over \(\binom{\mathcal{C}}{n_{\text{probes}}}\) subsets,
which is infeasible to compute. 

For results of FlashHead reported in the main text where the full log-likelihood is required, we estimate the true underlying distribution using a Monte-Carlo approximation, which we explain in more detail in Appendix \ref{app:mc}.

All hyper-parameters (number of clusters, number of probes, vocabulary size, window size, etc.) are identical to those listed in Section \ref{sec:experiments} of the main text. Datasets and their settings used in this section are the same as those originally described in Section \ref{subsec:lm-eval-tasks}. To ensure good precision, we accumulate probabilities in 64-bit floating point format (everything else remains in bfloat16).

\subsection{Monte-Carlo Simulation}
\label{app:mc}

For each of a total of \(N\) probe sets \(S_{1},\dots,S_{N}\sim p(S\mid\mathbf h)\) we evaluate the full conditional distribution \(p(v\mid\mathbf h,S_i)\) and accumulate it

\[
  \hat{s}_{N}(v\mid\mathbf h)\;=\;\sum_{i=1}^{N} p\!\bigl(v\mid\mathbf h,S_i\bigr).
\]

Averaging these \(N\) conditional distributions yields a properly normalised marginal,

\[
  \hat{p}_{N}(v\mid\mathbf h)
  \;=\;
  \frac{\hat{s}_{N}(v\mid\mathbf h)}{N}
  \;=\;
  \frac{1}{N}\sum_{i=1}^{N} p\!\bigl(v\mid\mathbf h,S_i\bigr),
  \qquad
  \sum_{t\in\mathcal V}\hat{p}_{N}(v\mid\mathbf h)=1.
\]

Because each probe set contributes a full distribution rather than a single token, this Monte-Carlo estimator has substantially lower variance than sampling individual tokens.

If a token is never present in any of the \(S_i\), then \(\hat{p}_{N}(v\mid\mathbf h)=0\) and the log-likelihood becomes \(-\infty\).  To avoid undefined accuracies we clip such zero entries to the minimum non-zero probability observed in the same context before computing task metrics.

\begin{figure}[H]
    \centering    \includegraphics[width=0.8\linewidth]{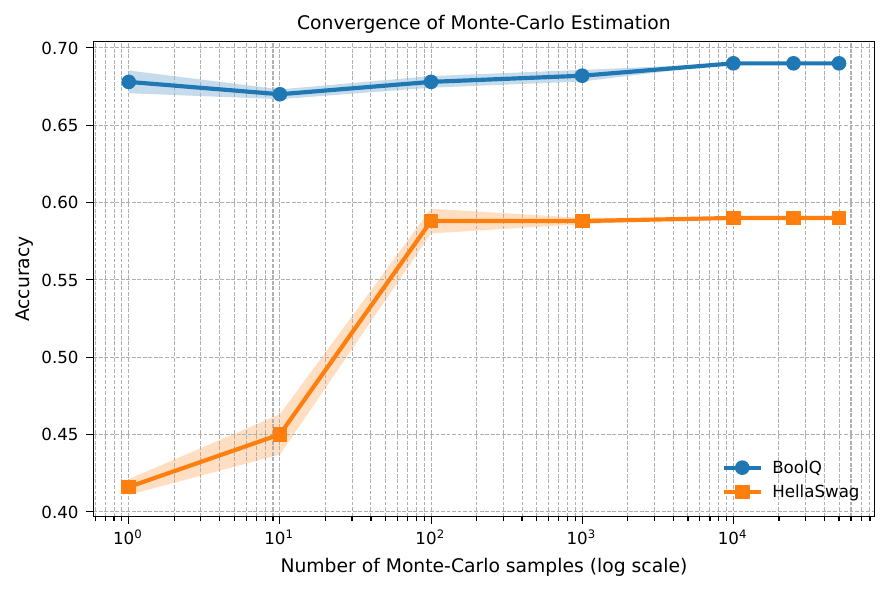}
    \caption{Convergence of the Monte-Carlo estimator when used to evaluate accuracy on the BoolQ and HellaSwag datasets. An average is taken over five independent clustering procedures with different random seeds. Shaded ribbons show $\pm1$ s.e.}
    \label{fig:mc_convergence}
\end{figure}

Figure~\ref{fig:mc_convergence} shows the accuracy obtained on BoolQ and HellaSwag as a function of \(N\). Beyond \(N\approx10{,}000\) the curves are flat, so we used that setting for all headline results.  With smaller sample sizes (\(<10{,}000\)) the “never-sampled‐token” issue becomes more common.  All Monte-Carlo runs were performed on an NVIDIA RTX 3500 Ada Geneneration; generating the 10 000 probe sets for each evaluation subset of HellaSwag or BoolQ took roughly 2 seconds.

\section{Models, hyper-parameters, datasets and evaluation settings}
In our experiments, we explore all methods using the baseline Llama-3.2-1B-Instruct model. The alternative heads are initialized with their own set of hyper-parameters (e.g., number of clusters, vocabulary size). We list all such settings in Table~\ref{appendix:models}.

To evaluate model performance (LM-Eval benchmark tasks and Top-K Containment) and on-device latencies, we use several datasets listed in Table~\ref{appendix:datasets}. For each dataset we provide both a formal literature reference and direct links to their public HuggingFace or GitHub repositories. In case of Top-K Containment, we use the Alpaca (eval) dataset and the test splits for the MATH-Hard and XNLI datasets. In the case of vocabulary trimming, we use the train split to mine the high frequency tokens to use for the reduced vocabulary.  

All LM-Eval benchmarks are conducted through the \href{https://github.com/EleutherAI/lm-evaluation-harness}{LM-Evaluation-Harness framework}. Table~\ref{appendix:eval-settings} details, for each benchmark, the reported metric and the relevant generation settings.

\begin{table}[ht]
  \centering
  \caption{Methods and hyper-parameters used in the various implementations.}
  \label{appendix:models}
  \begin{tabularx}{\textwidth}{l X}
    \toprule
    \textbf{Method} & \textbf{Hyper-parameters} \\
    \midrule
    \textbf{Baseline (LLaMA-3.2 1B Instruct)} &
    num\_hidden\_layers: 16 \\
    & hidden\_size: 2048 \\
    & intermediate\_size: 8192 \\
    & num\_attention\_heads: 32 \\
    & num\_key\_value\_heads: 8 \\
    & head\_dim: 64 \\
    & hidden\_act: silu \\
    & mlp\_bias: false \\
    & attention\_bias: false \\
    & rms\_norm\_eps: 1e-5 \\
    & max\_position\_embeddings: 131072 \\
    & rope\_scaling: \{factor: 32.0, high\_freq: 4.0 \\
    & \hspace{1em} low\_freq: 1.0, original\_max: 8192\} \\
    & rope\_theta: 500000.0 \\
    & bos\_token\_id: 128000 \\
    & eos\_token\_id: [128001, 128008, 128009] \\
    & vocab\_size: 128256 \\
    & dtype: bfloat16 \\
    \addlinespace
    \textbf{Vocab. Trimming} & vocab\_size: 64,000 \\
    & Top tokens filtered by frequency based on the Alpaca (train) dataset for all LM-Eval tasks, and the relevant train dataset for Top-K Containment evaluations \\
    \addlinespace
    \textbf{SVDSoftmax} & window: 256 \\
    & top\_n: 12,000 \\
    \addlinespace
    \textbf{Spherical K-Means} & n\_clusters: 100, 20. \\
    \addlinespace
    \textbf{Fast Graph Decoder (FGD)} & K: 384 \\
        & ef\_search: 300 \\
        & index\_M: 40 \\
        & ef\_construction: 300 \\
        & FGD implementation is CPU-only with dtype: float32 \\
    \midrule
  \end{tabularx}
\end{table}

\begin{table}[ht]
  \centering
  \caption{Overview of the datasets used in this paper.}
  \label{appendix:datasets}
  \begin{tabularx}{\textwidth}{l X X}
    \toprule
    \textbf{Dataset name} & \textbf{Reference} & \textbf{Link} \\
    \midrule
    Alpaca (train and eval) & ~\cite{alpaca2023, dubois2023alpacaeval} & \href{https://huggingface.co/datasets/tatsu-lab/alpaca}{tatsu-lab/alpaca}, \href{https://huggingface.co/datasets/tatsu-lab/alpaca_eval}{tatsu-lab/alpaca\_eval} \\
    MATH-Hard & ~\cite{hendrycksmath2021} & \href{https://huggingface.co/datasets/lighteval/MATH-Hard}{lighteval/MATH-Hard} \\
    XNLI & ~\cite{conneau2018xnli} & \href{https://huggingface.co/datasets/facebook/xnli}{facebook/xnli} \\
    MMLU-Pro & ~\cite{wang2024mmlupro} & \href{https://huggingface.co/datasets/TIGER-Lab/MMLU-Pro}{TIGER-Lab/MMLU-Pro} \\ 
    Hella-Swag & ~\cite{zellers-etal-2019-hellaswag} & \href{https://huggingface.co/datasets/Rowan/hellaswag}{hellaswag} \\
    IFEval & ~\cite{zhou2023instruction} & \href{https://huggingface.co/datasets/google/IFEval}{google/IFEval} \\
    BoolQ & ~\cite{clark2019boolq} & \href{https://huggingface.co/datasets/google/boolq}{google/boolq} \\
    BBH & ~\cite{suzgun2022challenging} & \href{https://github.com/suzgunmirac/BIG-Bench-Hard}{suzgunmirac/BIG-Bench-Hard}\\
    Truthful-QA (gen) & ~\cite{lin-etal-2022-truthfulqa} & \href{https://github.com/sylinrl/TruthfulQA}{sylinrl/TruthfulQA} \\
    OBQA & ~\cite{OpenBookQA2018} & \href{https://huggingface.co/datasets/allenai/openbookqa}{allenai/openbookqa} \\
    \midrule
  \end{tabularx}
\end{table}

\begin{table}[ht]
  \centering
  \caption{Evaluation settings for \href{https://github.com/EleutherAI/lm-evaluation-harness}{LM-Eval} tasks.}
  \label{appendix:eval-settings}
  \begin{tabularx}{\textwidth}{l X X}
    \toprule
    \textbf{Benchmark} & \textbf{Reported metric} & \textbf{Settings} \\
    \midrule
    MMLU-Pro & exact\_match,custom-extract & max\_gen\_toks:~2048 \\ && num\_fewshot:~5 \\ 
    HellaSwag & acc\_norm & num\_fewshot:~0 \\
    IFEval  & average of (prompt\_strict, instruction\_strict, prompt\_loose, instruction\_loose acc) & num\_fewshot:~0\\ && max\_gen\_toks:~1280 \\
    BoolQ & acc & num\_fewshot:~0 \\
    BBH & exact\_match,get\_answer & num\_fewshot:~3 \\ && max\_gen\_toks:~1024 \\
    Truthful-QA (gen) & bleu\_acc & num\_fewshot:~0 \\
    OpenBookQA (OBQA) & acc\_norm & num\_fewshot:~0\\
    \bottomrule
  \end{tabularx}
\end{table}

\section{Qualitative Comparison to Other Methods}

To qualitatively demonstrate the output of the different heads, one prompt from each of the three data sets Alpaca, MATH-Hard, and XNLI is listed in Table~\ref{tab:alpaca-qual}, \ref{tab:math-qual}, and \ref{tab:xnli-qual}.

\begin{table*}[ht]
\caption{Example outputs from the AlpacaEval dataset.}%
\label{tab:alpaca-qual}

\centering
\begin{adjustbox}{width=1.0\textwidth,center}
\begin{tabularx}{1.25\textwidth}{p{0.18\textwidth}X}
\toprule
\multicolumn{2}{p{\linewidth}}{\textbf{Prompt:} How did US states get their names?} \\[4pt]
\textbf{Method} & \textbf{Answer} \\
\midrule

Baseline &
\begin{minipage}[t]{\linewidth}\footnotesize\setlength{\parskip}{2pt}\raggedright
The origin of state names is a fascinating topic that spans centuries. Here's a brief overview.

\textbf{Early Names (1600s-1700s)}  
During the colonial era, many US states were named after:
1. \textbf{Native American tribes}: Many states were named after Native American tribes, such as Massachusetts (Massachusetts), New Hampshire (New Hampshire), and Pennsylvania (Pennsylvania).
2. \textbf{Colonial cities}: Some states were named after colonial cities, like Virginia (named after the English city of Virginia) and New York (named after the Dutch city of New Amsterdam).
3. \textbf{Geographic features}: States were also named after geographic features, such as Connecticut (Connecticut River) and Maryland (Maryland Peninsula).
\end{minipage}
\\
\addlinespace

FlashHead &
\begin{minipage}[t]{\linewidth}\footnotesize\setlength{\parskip}{2pt}\raggedright
The origin of the names of US states can be traced back to the early colonial period, when they were first established as territories, colonies, or provinces. Here's a brief overview of how each state got its name:
1. \textbf{Delaware}: Named after the Delaware River, which was named after the Delaware\ldots
\end{minipage}
\\
\addlinespace

Vocab. Trimming & {\footnotesize The origin of the names of US states is often traced back to the early colonial period, when European explorers and settlers arrived in the New World. Over time, the names of the states evolved through a combination of historical events, geographical features, and cultural influences. Here are some…} \\
\addlinespace

SVDSoftmax & {\footnotesize The names of US states were often chosen by their founders, and the process varied across states and over time. \textbf{New England States (1620s-1700s-present day states)} * Massachusetts (1620) - named after the Massachusetts Bay Colony * New Hampshire…} \\
\addlinespace

FGD & {\footnotesize The US has 50 different US, but there are actually 50 different "official" US names. The US has 50 different "alb" or "alb" (Alb) "alb" or "alb" "alb" "alb" or "alb" "alb" "alb" "alb" "alb" "alb" or "alb" "alb" "alb" "alb" "alb" "alb" "alb" "alb" "alb" "alb" "alb" "alb" "alb" "alb" "alb" "alb"…} \\
\addlinespace

\bottomrule
\end{tabularx}
\end{adjustbox}
\end{table*}

\begin{table*}[ht]
\caption{Example outputs from the MATH-Hard dataset.}%
\label{tab:math-qual}

\centering
\begin{adjustbox}{width=1.0\textwidth,center}
\begin{tabularx}{1.25\textwidth}{p{0.18\textwidth}X}
\toprule
\multicolumn{2}{p{\linewidth}}{\textbf{Prompt:} When Lauren was born on January 1, 1990, her grandparents put $\$1000$ in a savings account in her name. The account earned $7.5\%$ annual interest compounded quarterly every three months. To the nearest dollar, how much money was in her account when she turned two?} \\[4pt]
\textbf{Method} & \textbf{Answer} \\
\midrule

Baseline &
\begin{minipage}[t]{\linewidth}\footnotesize\setlength{\parskip}{2pt}\raggedright
We will use the formula for compound interest to solve this problem.

\#\# Step 1: Determine the interest rate per period  
The annual interest rate is $7.5\%$, and it is compounded quarterly. Since there are 4 quarters in a year, the interest rate per period is $7.5\%/4 = 1.875\%$.

\#\# Step 2: Calculate the number of periods  
Since the interest is compounded quarterly, the number of periods is 2 (1 quarter per year * 2 years = 2 periods).

\#\# Step 3: Calculate the interest earned  
The interest earned is the product of the principal, interest rate per period, and the number of periods. Interest $= \$1000 \times 1.875\% \times 2 = \$1000 \times 0.01875 \times 2 = \$37.50$.

\#\# Step 4: Calculate the total amount in the account  
The total amount in the account is the principal plus the interest earned. Total amount $= \$1000 + \$37.50 = \$1037.50$.
\end{minipage}
\\
\addlinespace

FlashHead &
\begin{minipage}[t]{\linewidth}\footnotesize\setlength{\parskip}{2pt}\raggedright
We will use the formula for compound interest to solve this problem.

\#\# Step 1: Determine the interest rate per period  
The annual interest rate is $7.5\%$, and it is compounded quarterly. Since there are 4 quarters in a year, the interest rate per period is $7.5\%/4 = 1.875\%$.

\#\# Step 2: Calculate the number of periods  
Since the interest is compounded quarterly, the number of periods is 2 (1 quarter per year * 2 years = 2 periods).

\#\# Step 3: Calculate the interest earned  
The interest earned is the product of the principal, interest rate per period, and the number of periods. Interest $= \$1000 \times 1.875\% \times 2 = \$1000 \times 0.01875 \times 2 = \$37.50$.

\#\# Step 4: Calculate the total amount in the account  
The total amount in the account is the principal plus the interest earned. Total amount $= \$1000 + \$37.50 = \$1037.50$.
\end{minipage}
\\
\addlinespace

Vocab. Trimming & {\footnotesize We assume that the interest is compounded $200$ times a year. \#\# Step 1: First, we need to understand the problem and the formula for compound interest. Compound interest is calculated using the formula $A = P(1 + \frac{r}{n})^n$, where $A$ is the amount of money accumulated after $n$ periods,…} \\
\addlinespace

SVDSoftmax & {\footnotesize We'll assume that the interest rate is compounded quarterly. \#\# Step 1: Calculate the number of compounding periods Since the interest is compounded quarterly, there are 4 compounding periods per year. However since the interest is compounded every 3 months, we need to calculate the number of…} \\
\addlinespace

FGD & {\footnotesize The total amount in the first year was $A$, and the total amount in the following three years was $B$. $A+B = 2 \, 200. \, 0 \, 0 \, 0$ \#\# A \#\# B \#\# A + B = 2,200. 0. 0. 0. \#\# A = 1000 + (0. 7 5 0) 4 0 0 * 1 2 0 0 \#\# A = 1000 + 0. 19 5 0 0 \#\# A = 1000 + 19. 5 0 0 \#\# A = 1000 + 19. 5 0 \#\# A = 1000…} \\
\addlinespace

\bottomrule
\end{tabularx}
\end{adjustbox}
\end{table*}

\begin{table*}[ht]
\caption{Example outputs from the XNLI dataset.}%
\label{tab:xnli-qual}

\centering
\begin{adjustbox}{width=1.0\textwidth,center}
\begin{tabularx}{1.25\textwidth}{p{0.18\textwidth}X}
\toprule
\multicolumn{2}{p{\linewidth}}{\textbf{Prompt:} Determine la relación entre las siguientes oraciones.

Premisa: "Y él dijo: Mamá, estoy en casa."
Hipótesis: "Llamó a su madre tan pronto como el autobús escolar lo dejó."

Opciones: (A) Implicación  (B) Neutral  (C) Contradicción

Respuesta:} \\[4pt]
\textbf{Method} & \textbf{Answer} \\
\midrule

Baseline &
\begin{minipage}[t]{\linewidth}\footnotesize\setlength{\parskip}{2pt}\raggedright
(B) Neutral

Análisis: Esta pregunta requiere la aplicación de la lógica de inferencia, específicamente la capacidad de identificar la relación entre dos oraciones. La hipótesis se basa en la premisa, que establece que el autobús escolar lo dejó en casa. Sin embargo, la oración hipotética no proporciona información adicional sobre el estado de ánimo o la situación de la madre, lo que significa que no hay suficiente información para inferir la relación entre las dos oraciones. Por lo tanto, la respuesta correcta es Neutral.
\end{minipage}
\\
\addlinespace

FlashHead &
\begin{minipage}[t]{\linewidth}\footnotesize\setlength{\parskip}{2pt}\raggedright
(B) Neutral

Análisis: Esta pregunta requiere la aplicación de la lógica de inferencia, específicamente la capacidad de identificar la relación entre dos oraciones. La hipótesis se basa en la premisa, que establece que el autobús lo dejó en casa. Sin embargo, la oración hipotética no proporciona información adicional sobre el autobús, lo que significa que no hay suficiente información para inferir la relación entre las dos oraciones. Por lo tanto, la respuesta correcta es Neutral.
\end{minipage}
\\
\addlinespace

Vocab. Trimming & {\footnotesize (A) Implicación Análisis: La oración principal es una exclamación de sorpresa, lo que implica que la oración que se está evaluando no es una exclamación de sorpresa. La oración principal no está en la primera persona, por lo que no puede ser una exclamación de sorpresa. Por otro lado, la oración…} \\
\addlinespace

SVDSoftmax & {\footnotesize B Análisis: La oración de hipótesis tiene un enfoque de juicio, ya que se hace una suposición sobre lo que sucedió después de que el autobús escolar de su hijo saliera. La oración de hipótesis también tiene un tono de afirmación, ya que se afirma que su hijo llamó a su madre de inmediato después…} \\
\addlinespace

FGD & {\footnotesize (A) la orfandad Anal g: Se puede prejusar de la información, pero no se puede jusg. Se puede jusg de la información, pero no se puede prejusar. Se puede jusg de la información y se puede juse de la información. Se puede juse de la información y se puede juse de la información. Se puede juse de la…} \\
\addlinespace

\bottomrule
\end{tabularx}
\end{adjustbox}
\end{table*}

\section{Examples of Clusters}

To demonstrate the clustering used by FlashHead, we provide ten (out of the total 8,016) randomly selected clusters in Table~\ref{tab:sampled-clusters}. These examples illustrate how the tokenizer groups related spellings, capitalization variants, and domain-specific affixes.

\begingroup
\providecommand{\tok}[1]{\texttt{#1}}
\providecommand{\tokensep}{,\penalty0\ }
\newcolumntype{Y}{>{\RaggedRight\arraybackslash}X}
\begin{table}[ht]
\centering
\caption{Tokens for 10 randomly sampled clusters from LLaMA-3.2-1B-Instruct. Tokens are comma separated without extra padding; perceived leading spaces belong to the tokens themselves.}
\label{tab:sampled-clusters}
\begin{adjustbox}{width=\textwidth,center}
\begin{tabularx}{\textwidth}{p{0.17\textwidth} Y}
\toprule
\textbf{Cluster} & \textbf{Tokens} \\
\midrule
661 & \tok{ change}\tokensep\tok{change}\tokensep\tok{Change}\tokensep\tok{ Change}\tokensep\tok{\_change}\tokensep\tok{.change}\tokensep\tok{\_CHANGE}\tokensep\tok{CHANGE}\tokensep\tok{ CHANGE}\tokensep\tok{ Changing}\tokensep\tok{-change}\tokensep\tok{changing}\tokensep\tok{(change}\tokensep\tok{Changing}\tokensep\tok{\textbackslash{}tchange}\tokensep\tok{/change} \\
\addlinespace
1593 & \tok{ javax}\tokensep\tok{.junit}\tokensep\tok{ javafx}\tokensep\tok{ TestBed}\tokensep\tok{.slf}\tokensep\tok{.jupiter}\tokensep\tok{.testng}\tokensep\tok{.jboss}\tokensep\tok{.joda}\tokensep\tok{ NUnit}\tokensep\tok{JUnit}\tokensep\tok{.hamcrest}\tokensep\tok{\textbackslash{}TestCase}\tokensep\tok{/gtest}\tokensep\tok{ junit}\tokensep\tok{.assertj} \\
\addlinespace
3824 & \tok{ distr}\tokensep\tok{ district}\tokensep\tok{ distur}\tokensep\tok{ districts}\tokensep\tok{ disturbing}\tokensep\tok{ distress}\tokensep\tok{ disturb}\tokensep\tok{ distortion}\tokensep\tok{ distraction}\tokensep\tok{ disturbed}\tokensep\tok{ distorted}\tokensep\tok{ disturbance}\tokensep\tok{ distort}\tokensep\tok{ distressed}\tokensep\tok{\_district}\tokensep\tok{ disturbances} \\
\addlinespace
4336 & \tok{ UART}\tokensep\tok{USB}\tokensep\tok{ usb}\tokensep\tok{\_UART}\tokensep\tok{\_USB}\tokensep\tok{ PWM}\tokensep\tok{ USART}\tokensep\tok{ uart}\tokensep\tok{\_PWM}\tokensep\tok{\_usb}\tokensep\tok{UART}\tokensep\tok{\_uart}\tokensep\tok{ pwm}\tokensep\tok{\_USART}\tokensep\tok{\_pwm}\tokensep\tok{PWM} \\
\addlinespace
4389 & \tok{ exem}\tokensep\tok{ exempt}\tokensep\tok{ exemption}\tokensep\tok{nesty}\tokensep\tok{ exponential}\tokensep\tok{ pardon}\tokensep\tok{ exemptions}\tokensep\tok{exo}\tokensep\tok{ Amnesty}\tokensep\tok{ spree}\tokensep\tok{ ambush}\tokensep\tok{ amnesty}\tokensep\tok{ pard}\tokensep\tok{ Ames}\tokensep\tok{ impunity}\tokensep\tok{ exon} \\
\addlinespace
4584 & \tok{ done}\tokensep\tok{done}\tokensep\tok{Done}\tokensep\tok{ dice}\tokensep\tok{\_done}\tokensep\tok{ Done}\tokensep\tok{.done}\tokensep\tok{(done}\tokensep\tok{\textbackslash{}tdone}\tokensep\tok{\_DONE}\tokensep\tok{ DONE}\tokensep\tok{ Gone}\tokensep\tok{.Done}\tokensep\tok{DONE}\tokensep\tok{ undone}\tokensep\tok{-done} \\
\addlinespace
5065 & \tok{ictureBox}\tokensep\tok{.ToolStrip}\tokensep\tok{ToolStrip}\tokensep\tok{.toolStrip}\tokensep\tok{ pictureBox}\tokensep\tok{.menuStrip}\tokensep\tok{ toolStrip}\tokensep\tok{.ToolStripButton}\tokensep\tok{pictureBox}\tokensep\tok{ ToolStrip}\tokensep\tok{.toolStripSeparator}\tokensep\tok{toolStrip}\tokensep\tok{.toolStripButton}\tokensep\tok{ Bunifu}\tokensep\tok{PictureBox}\tokensep\tok{ PictureBox} \\
\addlinespace
5251 & \tok{ pack}\tokensep\tok{pack}\tokensep\tok{ Pack}\tokensep\tok{PACK}\tokensep\tok{.pack}\tokensep\tok{\_PACK}\tokensep\tok{ packs}\tokensep\tok{Pack}\tokensep\tok{\_pack}\tokensep\tok{ PACK}\tokensep\tok{ pak}\tokensep\tok{ Packers}\tokensep\tok{-pack}\tokensep\tok{-Pack}\tokensep\tok{ Packs}\tokensep\tok{(pack} \\
\addlinespace
7376 & \tok{peed}\tokensep\tok{ acceler}\tokensep\tok{ acceleration}\tokensep\tok{ hast}\tokensep\tok{ accelerate}\tokensep\tok{accel}\tokensep\tok{ Acceler}\tokensep\tok{ accelerated}\tokensep\tok{\_accel}\tokensep\tok{Acceler}\tokensep\tok{ accelerator}\tokensep\tok{ accel}\tokensep\tok{ accelerating}\tokensep\tok{.accel}\tokensep\tok{Acceleration}\tokensep\tok{ accelerometer} \\
\addlinespace
7656 & \tok{ check}\tokensep\tok{check}\tokensep\tok{Check}\tokensep\tok{ Check}\tokensep\tok{\_check}\tokensep\tok{.check}\tokensep\tok{ checks}\tokensep\tok{ checking}\tokensep\tok{ CHECK}\tokensep\tok{-check}\tokensep\tok{(check}\tokensep\tok{ Checks}\tokensep\tok{\textbackslash{}tcheck}\tokensep\tok{Checking}\tokensep\tok{ Checking}\tokensep\tok{checking} \\
\bottomrule
\end{tabularx}
\end{adjustbox}
\end{table}
\endgroup

\end{document}